\documentclass[fleqn,10pt]{wlscirep}
\usepackage[utf8]{inputenc}
\usepackage[T1]{fontenc}
\usepackage{graphicx}
\usepackage{authblk}
\usepackage{siunitx}
\usepackage{booktabs}
\usepackage{caption}
\usepackage{placeins}
\usepackage{float}
\usepackage{color,soul}
\usepackage{bm}
\usepackage{makecell}
\usepackage{xurl}
\usepackage{tabularx}   
\usepackage{multirow}
\usepackage{times}
\usepackage[title]{appendix}
\usepackage{amsmath}
\usepackage{amsfonts}
\usepackage{amssymb}
\usepackage{xcolor}
\usepackage{xr}
\usepackage{multirow}
\usepackage{booktabs}
\usepackage[table]{xcolor}
\usepackage{threeparttable}
\usepackage{graphicx}
\usepackage{booktabs,longtable,pdflscape,multirow,threeparttable,siunitx}

\hypersetup{
    colorlinks=true,
    linkcolor=blue!60!black,
    citecolor=blue!60!black,
    urlcolor=blue!60!black,
}

\usepackage{array}
\usepackage[super,sort&compress,numbers]{natbib}
\usepackage{hyperref}

\makeatletter

\newcommand*{\addFileDependency}[1]{
\typeout{(#1)}
\@addtofilelist{#1}
\IfFileExists{#1}{}{\typeout{No file #1.}}
}\makeatother

\title{LiDAR Intensity Reconstruction with Limited Real Data: An Architecture Benchmark under Fixed-Epoch Training}

\author[1]{Christopher M. Coovrey}
\author[1]{Aniruddha Bora}
\affil[1]{Department of Computer Science, Texas State University, San Marcos, TX 78666, USA}

\begin{abstract}
Physics-derived simulation can supplement scarce real-world LiDAR intensity measurements, but its effectiveness depends on the learning architecture and the amount of real training data available. We benchmark nine neural architectures for LiDAR intensity reconstruction from depth, semantic labels, incidence angle, and material reflectance, using real SemanticKITTI and synthetic VoxelScape range-image data. All models follow a fixed 20-epoch protocol; this controls the number of training passes, not the computational cost across architectures. We report mean absolute error (MAE), mean squared error (MSE), peak signal-to-noise ratio (PSNR), structural similarity (SSIM), inference timing, and spectral-energy agreement. We also evaluate four selected architectures with fixed-size training mixtures containing 0\%, 1\%, 5\%, 10\%, 20\%, 50\%, or 100\% real frames. With real-only training, UNet-V2 achieves the best reported reconstruction metrics (MAE 0.234, MSE 0.140, PSNR 8.64 dB, and SSIM 0.647), followed by a denoising diffusion model (DDPM). In several mixtures containing 1--10\% real data, DDPM achieves lower error and higher image-quality metrics than the evaluated deterministic alternatives, whereas UNet-V2 leads with 100\% real data. Spectral analysis additionally reveals architecture-dependent discrepancies across frequency bands. These findings characterize reconstruction trade-offs under the stated data mixtures and fixed-epoch protocol, without establishing compute-matched superiority or isolating the diffusion objective from backbone differences.

\end{abstract}

\begin{document}
\flushbottom
\maketitle

\thispagestyle{empty}

\setcounter{secnumdepth}{0} 

\section{Introduction}
\label{section:Introduction}

LiDAR (light detection and ranging) provides geometric measurements and return-intensity information for robotic perception and autonomous driving \citep{royo2019,behley2021}. In addition to range, the recorded return intensity depends on surface properties, incidence angle, distance, and characteristics of the sensing system \citep{anand2024,anand2025}. Reproducing realistic intensity is useful for sensor simulation and learning-based processing, but large collections of real measurements and associated scene annotations are costly to assemble. Simulation can supplement real-world observations and enable systematic evaluation of learning systems \citep{guillard2022}. However, a model trained on simulated observations may not behave similarly on real measurements because the two domains differ in appearance, noise, scene composition, and sensor response \citep{xiao2024}.

Recent physics-aware LiDAR pipelines represent point clouds as organized range images and derive depth (D), semantic-label (L), incidence-angle (I), and material-reflectance (M) modalities \citep{anand2024,anand2025}. These DLIM features provide physically motivated inputs for intensity prediction. Prior work studied UNet- and Pix2Pix-based reconstruction using related inputs \citep{anand2025}. An open empirical question is whether the architecture ranking changes when the real-data fraction becomes small and synthetic data comprise most of the training and validation set. This issue is distinct from performance on a fully synthetic test set: deployment-relevant transfer requires testing on held-out real-world measurements.

We address this question through a benchmark of nine architectures spanning convolutional encoder--decoders, adversarial models, a vector-quantized autoencoder, a denoising diffusion model, and a flow-matching model. The experiments first measure each architecture on held-out data from the same source as training. We then study four representative architectures on fixed-size mixtures of real SemanticKITTI and synthetic VoxelScape frames, with evaluation on the same held-out SemanticKITTI test set. The mixture sweep is designed to describe sensitivity to real-data availability, rather than to establish a causal advantage of synthetic augmentation over compute-matched real-only training.

Our contributions are threefold:
\begin{itemize}
    \item We compare nine architectures for DLIM-conditioned LiDAR intensity reconstruction under a common 20-epoch training protocol and report reconstruction and inference metrics on real and synthetic datasets.
    \item We quantify how the performance of four selected architectures changes across real-data fractions from 0\% to 100\%, using a fixed total mixture size and a common real-domain evaluation set.
    \item We complement image-level metrics with frequency-band spectral-energy analysis to describe differences in the scales of intensity variation represented by the predictions.
\end{itemize}

A fixed epoch count does not equalize GPU time, floating-point operations, optimizer steps for differently sized datasets, or convergence across model families. Accordingly, the results should be interpreted as an architecture comparison under the stated protocol, not as a compute-matched efficiency ranking. Section~\ref{section:Related Work} reviews the relevant foundations, Section~\ref{section:Deep Learning Architectures} describes the architectures, Section~\ref{section:Results} presents the results, Section~\ref{section:Discussion Limitations} discusses limitations, and Section~\ref{section:Conclusion} summarizes the findings.

\section{Related Work}
\label{section:Related Work}

\paragraph{Real and synthetic LiDAR data.}
SemanticKITTI provides real-world LiDAR point clouds and semantic annotations associated with KITTI odometry sequences \citep{behley2021}. VoxelScape provides simulated LiDAR scans and scene annotations \citep{saleh2023}. These resources support studies of learning with real and synthetic measurements, but their distributions should not be assumed equivalent. In particular, synthetic-to-real shifts can affect LiDAR perception performance \citep{xiao2024}. Our study focuses on regression of LiDAR return intensity rather than semantic segmentation, and evaluates mixed-source models on held-out real data.

\paragraph{Physics-derived inputs for intensity reconstruction.}
Projecting LiDAR points into structured two-dimensional range images enables the use of image-to-image neural networks \citep{anand2024,anand2025}. Previous work combined depth, incidence angle, material reflectance, and semantic information to provide physically motivated features for intensity prediction. In an idealized return model, reflected power depends on range, the orientation of the surface relative to the incoming beam, and material reflectance; practical measured intensity additionally depends on sensor processing and other factors \citep{anand2024,anand2025}. The material-related features in this line of work draw on sources including NASA ECOSTRESS \citep{Meerdink2019}. We refer to these inputs as \emph{physics-derived modalities}, rather than interpreting their use alone as a physics-constrained training objective.

\paragraph{Learning architectures and prior baseline.}
UNet-style encoder--decoders support dense spatial regression through multiscale representations and skip connections \citep{ronneberger2015}. Conditional adversarial approaches such as Pix2Pix combine a generator with a discriminator to encourage realistic output structure \citep{isola2017,goodfellow2014}. Anand et al.\ \citep{anand2025} previously studied UNet and Pix2Pix for LiDAR intensity inference with physics-derived modalities, reporting the MSE values reproduced in Table~\ref{anand-table}. The present work broadens the architecture set and evaluates data-mixture sensitivity, but the reported performance should not be interpreted as a strictly controlled improvement over the earlier study because configurations and training schedules differ.

\begin{table}[t]
\caption{Previously reported LiDAR intensity reconstruction MSE from \citep{anand2025}; experimental settings differ from the present benchmark.}
\label{anand-table}
\begin{center}
\begin{tabular}{llc}
\toprule
Architecture & Dataset & MSE \\
\midrule
\multirow{2}{*}{UNet} & SKITTI & 0.198 \\
 & VSCAPE & 0.231 \\
\midrule
\multirow{2}{*}{Pix2Pix} & SKITTI & 0.190 \\
 & VSCAPE & 0.174 \\
\bottomrule
\end{tabular}
\end{center}
\end{table}

\paragraph{Spectral diagnostics.}
The power spectrum describes the distribution of squared Fourier magnitude across spatial frequencies. Low-frequency components generally summarize broad intensity variation, while higher-frequency components capture finer variation and sharp transitions. Spectral diagnostics have been considered in studies of learning dynamics and physical-system reconstruction \citep{bordelon2020,khodarkarami2026,oommen2025,chakraborty2026}. Comparing predicted and target spectra can expose differences not visible in aggregate pixel-error metrics. However, spectral-energy agreement alone does not establish spatial alignment because the power spectrum does not retain Fourier phase. We therefore treat frequency-band errors as complementary descriptive diagnostics, not as proof of object-level reconstruction or generalized physical understanding.

\paragraph{Scope of the present benchmark.}
We compare nine implementations with roughly similar reported model parameter counts and a shared 20-epoch cap. In addition to MAE and MSE, we report PSNR, SSIM, and inference timing. We then examine mixtures of SemanticKITTI and VoxelScape frames on a fixed held-out real test set. The principal contribution is the observed dependence of performance on model choice and real-data fraction within this specific protocol.

\section{Deep Learning Architectures}
\label{section:Deep Learning Architectures}

We evaluate nine implemented architectures spanning direct regression, adversarial learning, discrete latent modeling, diffusion, and flow matching. Table~\ref{arch-table} summarizes their reported configurations. Parameter counts are approximately 7.6--8.2 million for the listed networks, but similar parameter counts do not imply equivalent training or inference cost. Moreover, Pix2Pix and RaGAN are listed by generator parameter count and also employ discriminators during training; direct regressors and generative models use different objectives. All architectures follow the same 20-epoch cap, with model-specific learning rates and training procedures.

\begin{table}[t]
\begin{threeparttable}
\caption{Reported architecture configurations under the fixed 20-epoch protocol}
\label{arch-table}
\begin{center}
\begin{tabular}{lccccccc}
\toprule
Architecture & Enc & Dec & Params & Cond & Loss & LR & WD \\ \hline
UNet-V1 & 256, 128, 64 & 64, 128, 156 & 7.7M & F & $L^2$ & 5e-4 & 1e-5 \\ \hline
Pix2Pix (Gen) & 352, 176, 88, 44, 22 & 22, 44, 88, 176, 352 & 7.9M & T & $L^1$ & 1e-4 & 1e-5 \\ \hline
UNet-V2 & 256, 128, 64 & 64, 128, 156 & 7.7M & F & $L^2$ & 5e-4 & 1e-5 \\ \hline
RaGAN (Gen) & 352, 176, 88, 44, 22 & 22, 44, 88, 176, 352 & 7.9M & T & $L^1$, Rel-Avg & 1e-4 & 1e-5 \\ \hline
VQ-VAE2 & 256, 32, 16, 8 & 8, 16,32, 256 & 7.6M & F & $L^2$ & 5e-4 & 1e-5 \\ \hline
IUNet & 288, 144, 72 & 72, 144, 288 & 7.9M & F & $L^2$ & 5e-4 & 1e-5 \\ \hline
M\&M & 280, 140, 70 & 70, 140, 280 & 7.8M & F & $L^2$ & 5e-4 & 1e-5 \\ \hline
DDPM & 232, 116, 58 & 58, 116, 232 & 8.2M & T & $L^2$ & 5e-5 & 1e-5 \\ \hline
FlwMtch & 232, 116, 58 & 58, 116, 232 & 8.2M & T & $L^2$ & 5e-5 & 1e-5 \\ \hline
\bottomrule
\end{tabular}
\begin{tablenotes}
    \item Enc: reported encoder channel widths
    \item Dec: reported decoder channel widths
    \item Params: reported parameter count (generator only for Pix2Pix and RaGAN)
    \item Cond: reported explicit conditioning flag (T/F); all architectures use DLIM inputs
    \item Loss: reported principal loss term (generative methods may have additional terms)
    \item LR: Learning Rate
    \item WD: Weight Decay
\end{tablenotes}    
\end{center}
\end{threeparttable}
\end{table}

\textbf{UNet-V1} follows the foundational UNet architecture defined by \citep{ronneberger2015} and applied in the context of LiDAR prediction by \citep{anand2025}. This architecture features three encoder and three decoder levels connected via skip connections and max pooling operations. It utilizes standard convolution blocks composed of 2D convolutions, BatchNorm, ReLU and dropout layers to process spatial representations effectively.

\textbf{Pix2Pix} is based on the conditional GAN framework introduced by \citep{isola2017} and is also 
utilized in the work of \citep{anand2025}. This model consists of a coupled generator and discriminator
network. The generator employs five encoder and five decoder levels built with convolution blocks
containing 2D convolutions, BatchNorm and LeakyReLU activations, concluding with an output layer that 
contains convolution and average pooling. For $L^1$ training loss, Pix2Pix has $\beta$ of (0.50, 0.999), $L^1 \lambda$ of 100, and $L^1$ gradient penalty of 10. 

\textbf{UNet-V2} is a variant of UNet-V1 with targeted changes to activation and regularization. While 
maintaining a similar structural layout as UNet-V1, it modifies the encoder and decoder convolution blocks by 
replacing ReLU activations with GELU and removing intermediate dropout layers, while retaining dropout
exclusively within the network bottleneck.

\textbf{RaGAN} adapts the Pix2Pix conditional GAN framework by incorporating a relativistic
average discriminator \citep{jolicoeur-martineau2018} (Rel-Avg). By evaluating the relative realism of generating data versus real data during training, this architecture modifies the standard adversarial training loss to improve convergence and quality. For $L^1$ training loss, RaGAN has $\beta$ of (0.50, 0.999), $L^1 \lambda$ of 100 without gradient penalty. 

\textbf{VQ-VAE2} is a hierarchical vector quantized variational autoencoder based on the architecture
detailed by \citep{razavi2019}. It leverages a structured multi-scale design featuring four bottom-up
encoders and four top-down decoders, enabling the generation of diverse, high-fidelity images through
discrete latent representations. 

\textbf{IUNet} incorporates Inception-style multibranch convolutions motivated by \citep{szegedy2015,szegedy2016} and the implementation described by \citep{bora2025}. IUNet consists of three encoders and three decoders that feature specialized inception style multi-branch convolutions, BatchNorm, and GELU activations. 

\textbf{M\&M} is a multi-scale architecture defined by \citep{bora2025}. It consists of three specialized encoders that fuse strided convolutions, max pooling, and average pooling operations. Alongside the encoders, there are three decoders equipped with skip connections that combine transposed convolutions, inception upsampling, and pixel shuffle techniques.

\textbf{DDPM} (Denoising Diffusion Probabilistic Models), as formalized by \citep{ho2020}, utilizes an
encapsulated backbone architecture referred to here as UNet-V3. This backbone architecture introduced by \citep{oommen2025}, features three encoders and three decoders built with Residual Blocks that incorporate both timestep and feature information using 2D convolutions, RMSNorm, and SiLU activations. The diffusion framework uses a forward q-process to add white noise to the target and a training loss to learn the reversed noise process. DDPM uses conditioning and attention with sinusoidal position embedded sampling. 

\textbf{FlwMtch} (Flow Matching) utilizes the same UNet-V3 backbone architecture, but replaces the traditional diffusion q-process with an optimal transport flow matching process \citep{lipman2023} using $\sigma_{min}$ of 1e-4, conditioning, attention, and sinusoidal position embedded sampling. This supplies a different generative training objective while retaining the same reported backbone family. Because UNet-V3 differs from UNet-V2, comparisons between DDPM and UNet-V2 jointly reflect changes in backbone and objective.

\section{Results}
\label{section:Results}

We evaluate intensity reconstruction using real SemanticKITTI (SKITTI) and simulated VoxelScape (VSCAPE) scans represented as organized two-dimensional images. The SemanticKITTI split uses sequences 00--02 for training (4,541, 1,101, and 1,858 frames; 7,500 in total), 2,500 frames from sequence 08 for validation, and sequences 09--10 for testing (1,591 and 909 frames; 2,500 in total). The VoxelScape split contains 7,500 training frames selected from sequences 03--10, 2,500 validation frames selected from sequences 11--13, and 2,500 test frames selected from sequences 14--16. These splits separate the selected sequences across training, validation, and testing.

Depth, semantic-label, incidence-angle, and material-reflectance (DLIM) representations are derived for each included frame following the physics-based preprocessing described in prior work \citep{anand2024,anand2025}. The intensity targets are filtered using the procedure illustrated in Figure~\ref{fig:app_filter_target}; details and the reported threshold notation are discussed in Appendix A.1. We evaluate MAE, MSE, PSNR (data range 1.0), SSIM (window size 7; data range 1.0; uniform window), and the reported inference-time cost (ICOST). Image-level metrics and the spectral diagnostics address different aspects of reconstruction fidelity.

All nine architectures are trained under a common cap of 20 epochs, using the model-specific hyperparameters summarized in Table~\ref{arch-table}. This design fixes training passes through each corresponding training set but does not normalize GPU time, floating-point operations, training-step cost, or convergence. Comparisons of computational efficiency are therefore restricted to the separately reported inference timing; even those values depend on the implementation and timing protocol described in Appendix A.2.

For mixed-source training, a fixed-size set of 7,500 training frames is constructed from $X\%$ real SemanticKITTI and $(100-X)\%$ synthetic VoxelScape data, for $X\in\{0,1,5,10,20,50,100\}$. The corresponding validation mixture contains 2,500 frames, and every mixture is evaluated on the same held-out set of 2,500 real SemanticKITTI frames. Since the real-only comparisons in Appendix A.4 can use fewer total frames than the mixed-source runs, those comparisons do not control the number of example exposures or optimizer updates. We interpret them descriptively rather than as compute-matched demonstrations of synthetic-data benefit. We analyze spectral-energy discrepancies using Equation~\ref{eq:spec_ene_err} and the frequency bands described in Appendix A.6 and A.7.

\subsection{Architecture Performance Metrics}

We use mean absolute error (MAE) and mean squared error (MSE) to quantify per-pixel discrepancies, with MSE placing greater weight on large residuals. Peak signal-to-noise ratio (PSNR) transforms squared reconstruction error into a logarithmic scale given the stated image data range. Structural similarity (SSIM) compares local luminance, contrast, and structure. We additionally report inference cost (ICOST), as defined by the measurement convention in Appendix A.2. MAE and MSE are interpreted here as numerical errors on the normalized intensity representation, consistent with the stated unit data range for PSNR and SSIM. The implementation-level normalization and validity-mask conventions are not fully specified in the current manuscript. Finally, spectral-energy plots provide complementary information about the relative power of spatial-frequency components, but do not measure whether predicted objects occupy the correct pixel locations.

\begin{figure}[h]
\begin{center}
\fbox{%
\includegraphics[width=0.95\linewidth]{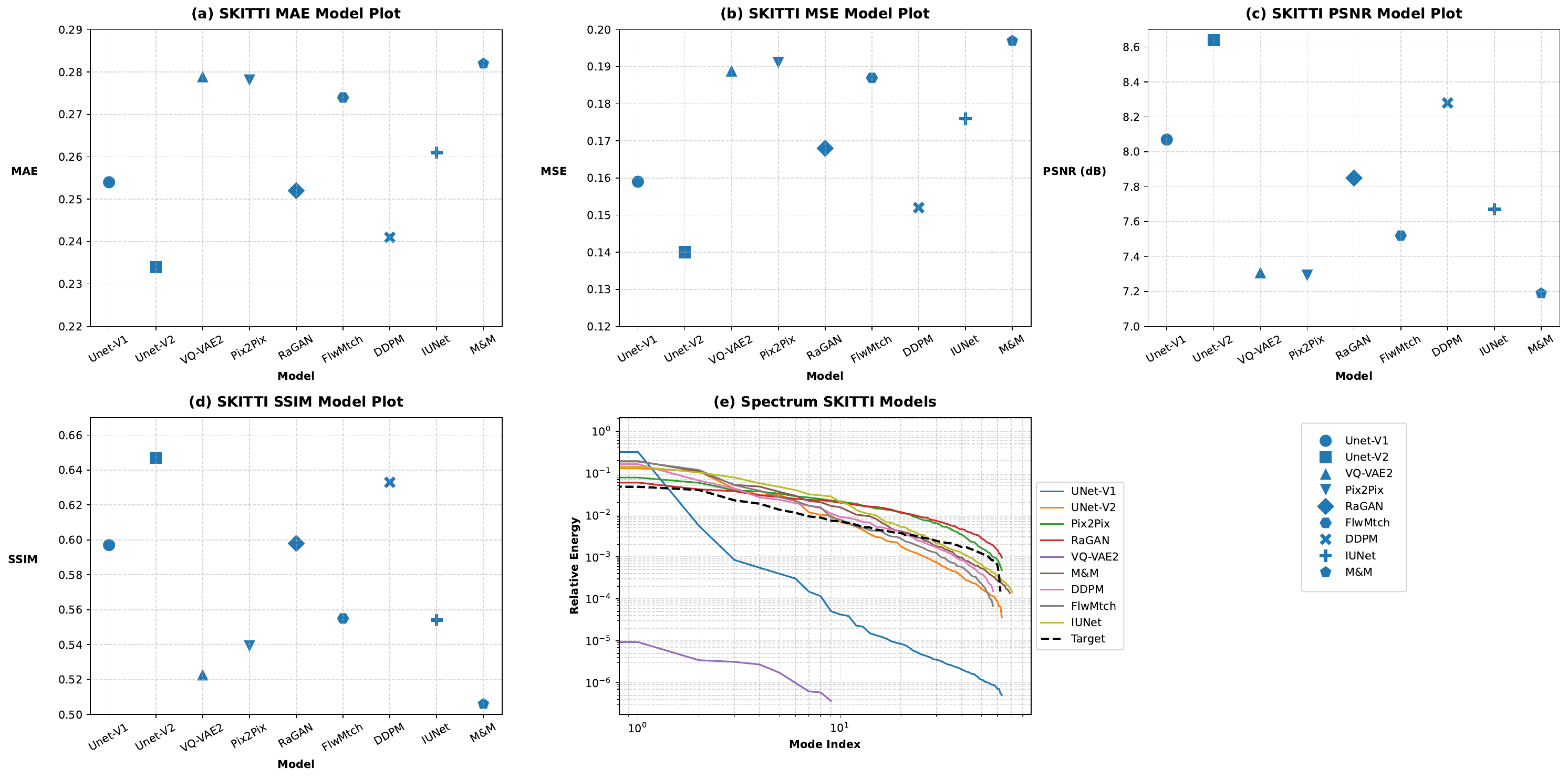}%
}
\end{center}
\caption{SKiTTI performance metrics.}
\label{fig:skitti_performance_metrics}
\end{figure}

Figure~\ref{fig:skitti_performance_metrics} summarizes the results for models trained and evaluated on SemanticKITTI; exact values appear in Table~\ref{performance-table}. UNet-V2 achieves the best reported performance on all four reconstruction metrics: MAE 0.234, MSE 0.140, PSNR 8.64 dB, and SSIM 0.647. DDPM ranks next on the same metrics (MAE 0.241, MSE 0.152, PSNR 8.28 dB, and SSIM 0.633). VQ-VAE2, Pix2Pix, flow matching, and M\&M have larger errors or lower structural-similarity scores under the 20-epoch protocol. These outcomes establish relative performance at the chosen training endpoint; they do not by themselves identify whether particular models are limited by architecture, optimization, or convergence.

Panel (e) compares the target and predicted intensity power spectra. Some architectures display substantially lower spectral power than the target at portions of the frequency range, while others retain comparatively greater power. Both deviations can coexist with poor image-level reconstruction. As phase information is not shown, the spectra should be interpreted as measurements of frequency-dependent energy alignment rather than direct evidence of spatial or semantic fidelity.

\begin{figure}[h]
\begin{center}
\fbox{%
\includegraphics[width=0.95\linewidth]{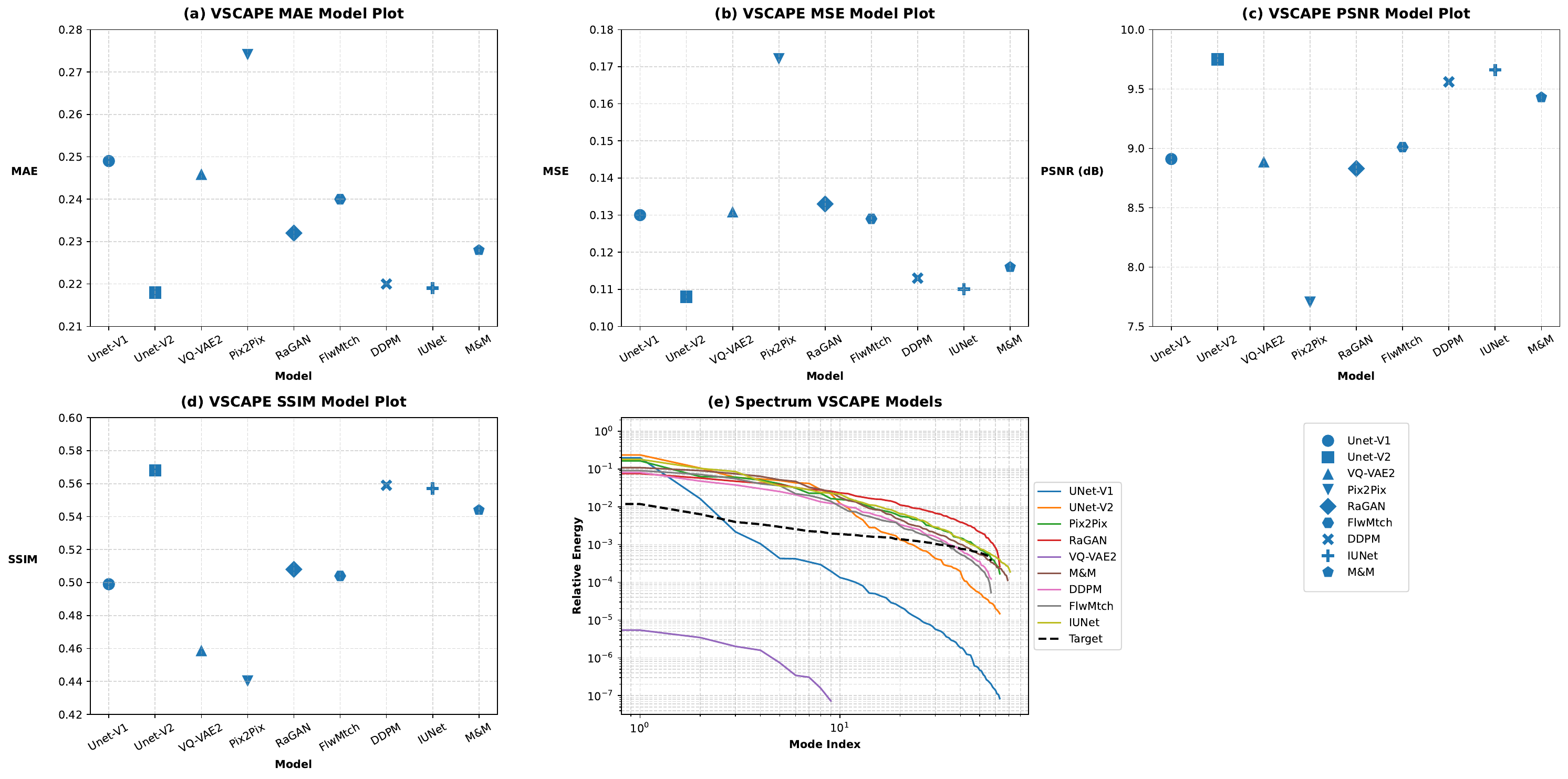}%
}
\end{center}
\caption{VSCAPE performance metrics.}
\label{fig:vscape_performance_metrics}
\end{figure}

Figure~\ref{fig:vscape_performance_metrics} shows the corresponding results for models trained and evaluated on VoxelScape, again using separate held-out sequences from that source. UNet-V2 obtains the lowest reported MSE (0.108), highest PSNR (9.75 dB), and highest SSIM (0.568), with MAE 0.218. IUNet (MAE 0.219, MSE 0.110, PSNR 9.66 dB, SSIM 0.557) and DDPM (MAE 0.220, MSE 0.113, PSNR 9.56 dB, SSIM 0.559) are close competitors; M\&M also performs better on VoxelScape than in the separate SemanticKITTI experiment. Pix2Pix has the largest VoxelScape MAE and MSE and the lowest SSIM among the reported models. The differences between the two datasets are consistent with architecture-dependent sensitivity to the evaluation domain, but because these are separate within-domain training-and-testing experiments, they should not themselves be interpreted as direct measurements of synthetic-to-real transfer.

\subsection{Synthetic Blending Architecture Performance}

We next study the effect of the real-data fraction within fixed-size training mixtures. Each model receives 7,500 training frames in total and is evaluated on the same held-out 2,500-frame SemanticKITTI test set. The real fractions are $X\in\{0,1,5,10,20,50,100\}\%$. For example, the 1\% mixture uses 75 real and 7,425 synthetic training frames, while the 100\% mixture uses 7,500 real frames. The matching validation mixtures contain 2,500 frames. The real and synthetic subsets for a given fraction are shared across the evaluated architectures, according to the experimental setup.

Four representative models are studied in detail: UNet-V2 and DDPM, which obtain the strongest performance on the fully real-data comparison, together with IUNet and M\&M, which exhibit comparatively large changes between real-trained and synthetic-trained evaluations in Appendix A.3. A large relative change denotes sensitivity to the training domain and does not necessarily denote a favorable transfer outcome. The mixture experiment measures how the selected models respond to different training compositions at a constant total mixture size; it does not isolate the impact of synthetic data from potential interactions with the available real examples.

\begin{figure}[h]
\begin{center}
\fbox{%
\includegraphics[width=0.95\linewidth]{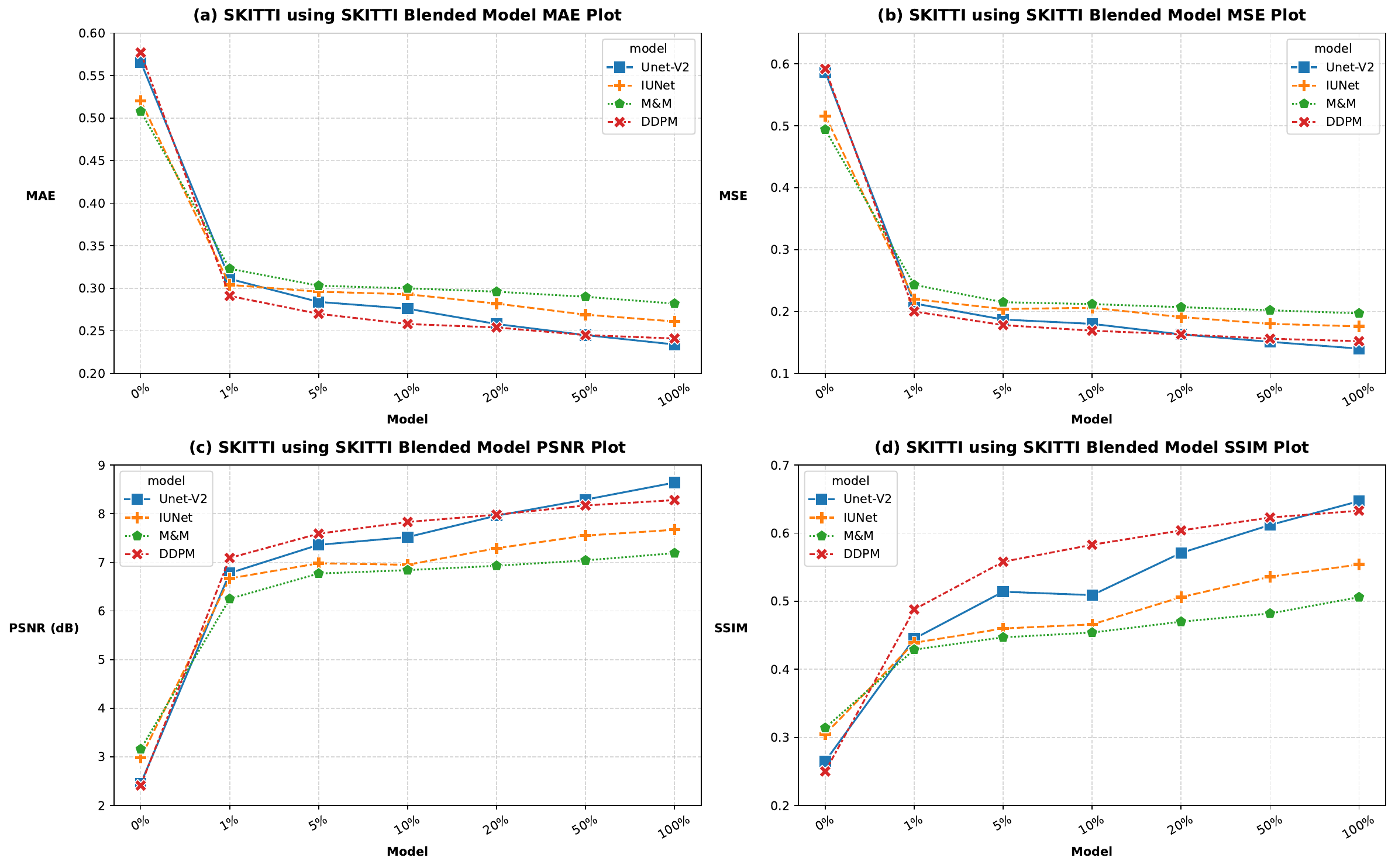}%
}
\end{center}
\caption{SKITTI blending performance.}
\label{fig:skitti_blend_metrics}
\end{figure}

Figure~\ref{fig:skitti_blend_metrics} reports reconstruction metrics across the selected mixtures. At 0\% real data, all four architectures exhibit comparatively poor performance on the real test set, illustrating the difficulty of direct synthetic-only transfer under the present setup. The inclusion of a small fraction of real training frames is associated with a marked improvement for every model. Across several mixtures with 1--10\% real data, DDPM has the lowest reported MAE and among the strongest MSE and PSNR values. Its SSIM exceeds those of the other selected architectures at the intermediate mixtures shown, but not at the 100\% real-data endpoint: there UNet-V2 achieves SSIM 0.647 compared with DDPM's 0.633. As the real-data fraction increases, UNet-V2 becomes increasingly competitive and achieves the best reported MAE, MSE, PSNR, and SSIM under 100\% real-data training.

IUNet and M\&M show different responses to composition and have relatively competitive synthetic-only results among the selected models on some metrics. Nonetheless, their absolute reconstruction metrics generally improve as real-data availability increases. The key result is thus a change in relative architectural ranking across mixture ratios, not an absolute deterioration of these models when real frames are introduced. Because the study contains only selected architectures in the mixture sweep, the findings are limited to that subset.

\begin{figure}[h]
\begin{center}
\fbox{%
\includegraphics[width=0.8\linewidth]{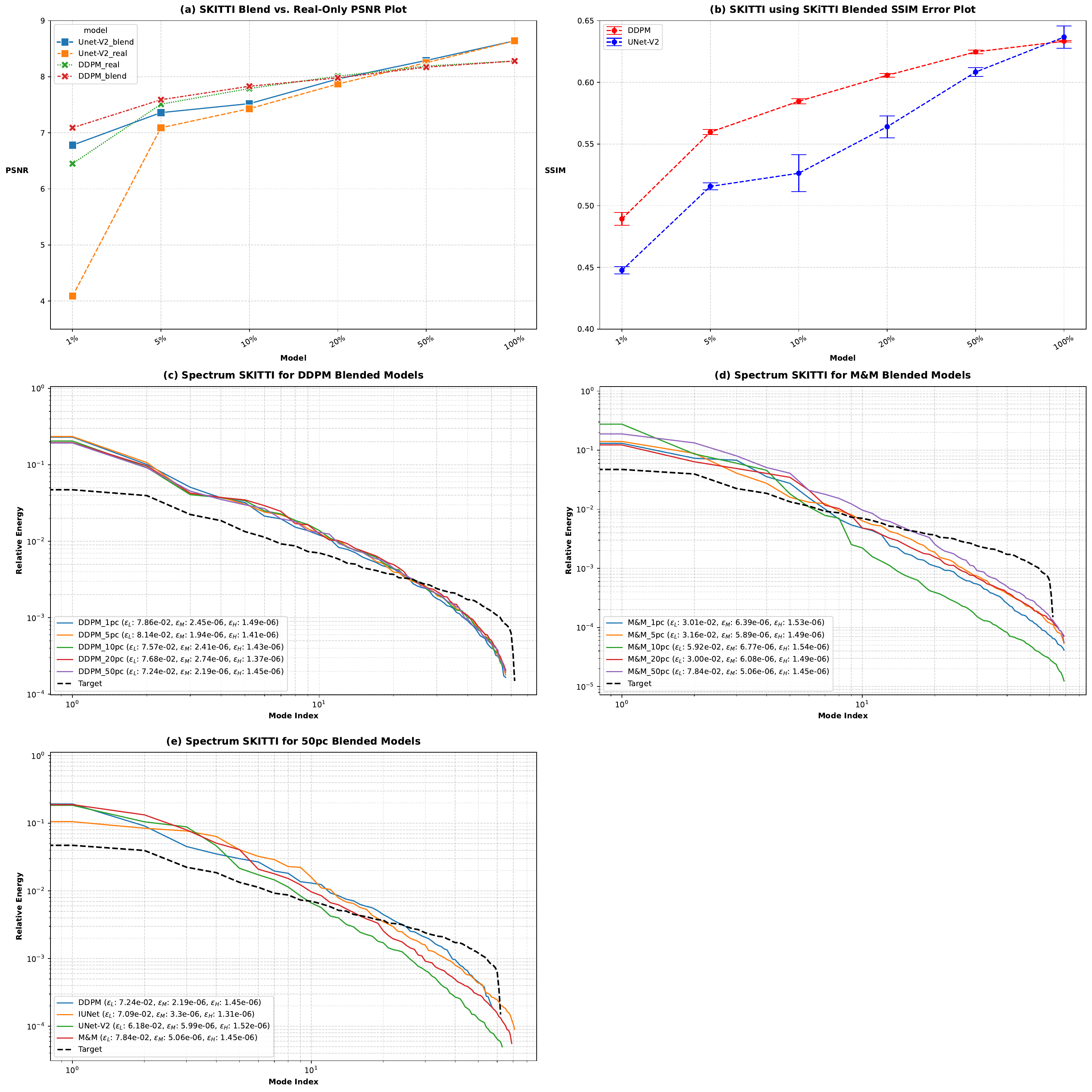}%
}
\end{center}
\caption{SKITTI blending ablations.}
\label{fig:skitti_blend_ablations}
\end{figure}

Figure~\ref{fig:skitti_blend_ablations} provides supplementary comparisons of blended and real-only training, SSIM variation, and spectral-energy agreement; fuller plots appear in Appendix A.4--A.7. Panel (a) shows that the gap between blended and real-only PSNR differs between UNet-V2 and DDPM, particularly at the 1\% real-data level. This is an informative descriptive result, but the real-only run at 1\% uses only 75 training frames, whereas the blended run uses 75 real plus 7,425 synthetic frames. Under the same 20-epoch cap, the total number of examples processed differs substantially, so the observed gap cannot be attributed solely to synthetic-data information. Panel (b), based on three random seeds for UNet-V2 and DDPM, indicates that the reported SSIM means and variability also depend on mixture proportion. With only three seeds, these intervals should be viewed as preliminary rather than conclusive evidence of ranking stability.

Panels (c)--(e) examine spectral-energy errors. DDPM frequently exhibits smaller mid-frequency errors, whereas M\&M achieves comparatively small low-frequency errors for several mixtures. The ranking is not uniform in every band and mixture: for example, Figure~\ref{fig:skitti_blend_ablations}(e) reports a lower high-frequency error for IUNet than for DDPM at 50\% real data. Accordingly, the evidence supports architecture-dependent frequency-band alignment rather than universal dominance of one model at all spatial scales. Moreover, similarity of Fourier magnitudes does not guarantee reconstruction of the correct object positions or boundaries.

\section{Discussion and Limitations}
\label{section:Discussion Limitations}

The joint results suggest a regime-dependent choice between architectures. UNet-V2 is a strong direct-regression baseline when all 7,500 real training frames are available, and its reported inference cost is lower than DDPM's. DDPM, by contrast, is competitive at several low real-data mixture fractions. This finding may motivate generative modeling in data-scarce scenarios, but the experiments do not establish that the diffusion learning objective is the sole source of the advantage. The DDPM implementation uses the UNet-V3 backbone, whereas UNet-V2 differs in residual structure, normalization, activation, and conditioning; backbone and objective effects are therefore confounded.

The results also distinguish within-domain reconstruction from transfer to real data. The models perform more strongly when trained and tested within the VoxelScape domain than when synthetic-only training is followed by evaluation on SemanticKITTI. Introducing a small number of real training examples is associated with large improvements in the latter setting. The reported curves characterize this behavior for the available sequences and mixture construction; they do not identify the specific dataset characteristics responsible for the domain difference. Similarly, DLIM inputs are physics-derived features, but none of the benchmarked objectives is shown here to enforce a LiDAR return equation or a physical constraint.

The benchmark has several limitations. First, the common 20-epoch cap is not a matched computational budget and may leave different architectures at different points in their optimization trajectories. Second, the blending study varies data composition at a fixed total of 7,500 training frames, while the real-only comparisons use smaller datasets at low real-data fractions; those settings differ in example exposure and potentially in update count. Third, the generative architectures are not compared against direct regression using the identical UNet-V3 backbone, preventing attribution of their performance differences solely to diffusion or flow-matching objectives. Fourth, the main evaluation uses a single real-world dataset with selected held-out sequences and a limited set of reported random seeds, which restricts conclusions about sensor-independent and statistically robust generalization.

Two additional interpretation limits concern the observables. The reported spectral-energy comparison discards phase and cannot establish spatial correctness on its own. Also, the experiments depend on DLIM features, including semantic and material-related information that may not be available without additional processing in deployment. Finally, the current tables report timing under an implementation-specific ICOST definition; without a fully specified timing protocol, they should not be treated as universal latency benchmarks. 

\section{Conclusion}
\label{section:Conclusion}

We benchmarked nine architectures for LiDAR intensity reconstruction from physics-derived depth, semantic-label, incidence-angle, and material-reflectance inputs using real SemanticKITTI and simulated VoxelScape data. Under the common 20-epoch training protocol, UNet-V2 achieves the strongest reported reconstruction metrics on the fully real-data comparison (MAE 0.234, MSE 0.140, PSNR 8.64 dB, SSIM 0.647), with DDPM the closest competitor (MAE 0.241, MSE 0.152, PSNR 8.28 dB, SSIM 0.633). IUNet and M\&M are relatively more competitive in the separate VoxelScape evaluation than in the fully real-data evaluation, highlighting that within-domain rankings depend on the source distribution.

In the mixed-source experiments, the evaluated models respond differently to the real-data fraction. DDPM performs strongly in several configurations with 1--10\% real training data, while UNet-V2 leads at 100\% real data. The accompanying spectral-energy diagnostics expose different patterns of frequency-band agreement, which supplement but do not replace pixel and structure-based fidelity metrics. Taken together, the results provide a descriptive benchmark of architecture choice under fixed-epoch training and changing real/synthetic composition. They do not establish matched-compute superiority, causal benefits of synthetic augmentation over real-only training, or a uniquely diffusion-driven mechanism. These distinctions are important for selecting models and for designing subsequent controlled studies of data-efficient LiDAR intensity prediction.

\clearpage
\bibliography{lidar_2026}
\clearpage

\appendix
\section{Additional Experimental Details}

\subsection{Intensity Target Filtering}

\begin{figure}[h]
\begin{center}
\fbox{%
\includegraphics[width=0.95\linewidth]{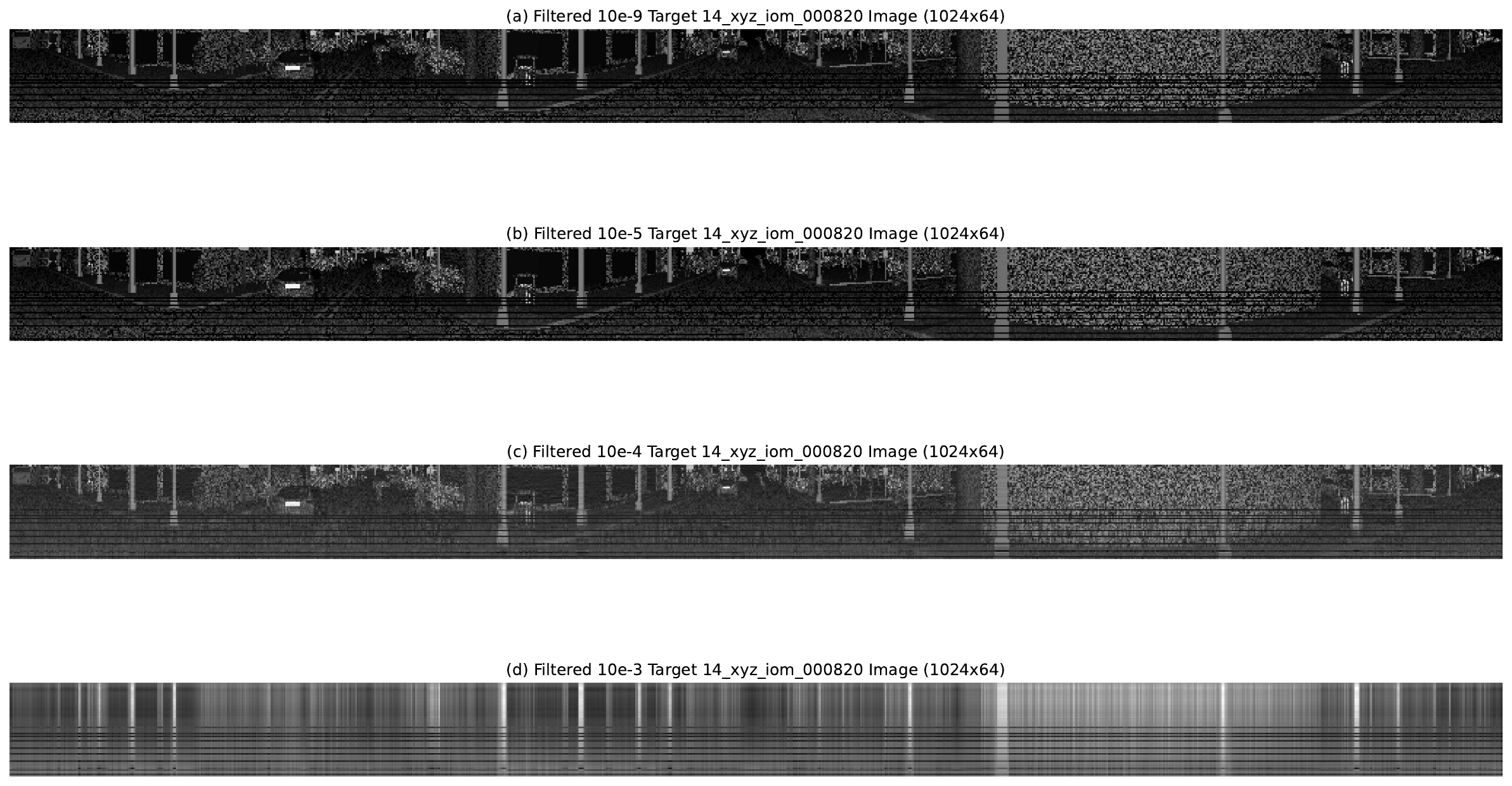}%
}
\end{center}
\caption{Example VoxelScape intensity targets under the reported filtering thresholds.}
\label{fig:app_filter_target}
\end{figure}

The authors inspected filtered VoxelScape intensity images to select a target-preprocessing threshold. Figure~\ref{fig:app_filter_target} displays one scene (14\_xyz\_iom\_000820) at several reported filtering settings. In the example, the weaker filtering settings preserve visible objects and road markings, whereas stronger filtering progressively suppresses these features. The source description specifies the selected filtering threshold using the literal notation \texttt{10e-5}, which is retained here to avoid silently changing the documented configuration. The precise numerical interpretation and mathematical definition of the preprocessing operation are not fully specified by the existing description. Visual inspection of one or more scenes motivates the preprocessing choice but does not by itself establish that the resulting targets preserve all physically meaningful weak returns.

\subsection{Model Performance Metrics for SKITTI and VSCAPE}
\label{section:app_perf_skitti_vscape}

Table~\ref{performance-table} gives the reported metrics for each architecture on both sources. MAE and MSE are presented as numerical errors on the normalized target scale, PSNR in dB, SSIM as a dimensionless index, and ICOST in seconds according to Equation~\eqref{eq:icost}. The precise intensity normalization and validity-mask conventions are not fully described in the present manuscript.

\begin{table}[t]
\caption{Reported LiDAR intensity reconstruction performance for separately trained and tested SemanticKITTI and VoxelScape models. Lower is better for MAE, MSE, and ICOST; higher is better for PSNR and SSIM.}
\label{performance-table}
\begin{center}
\begin{tabular}{lcccccc}
\toprule
Architecture & Dataset & MAE & MSE & PSNR (dB) & SSIM & ICOST (s) \\
\midrule
\multirow{2}{*}{UNet-V1} & SKITTI & 0.254 & 0.159 & 8.07 & 0.597 & 0.0061 \\
 & VSCAPE & 0.249 & 0.130 & 8.91 & 0.499 & 0.0061 \\
\midrule
\multirow{2}{*}{UNet-V2} & SKITTI & 0.234 & 0.140 & 8.64 & 0.647 & 0.006 \\
 & VSCAPE & 0.218 & 0.108 & 9.75 & 0.568 & 0.006 \\
\midrule
\multirow{2}{*}{VQ-VAE2} & SKITTI & 0.279 & 0.189 & 7.31 & 0.523 & 0.0085 \\
 & VSCAPE & 0.246 & 0.131 & 8.89 & 0.459 & 0.0085 \\
\midrule
\multirow{2}{*}{Pix2Pix} & SKITTI & 0.278 & 0.191 & 7.29 & 0.539 & 0.0054 \\
 & VSCAPE & 0.274 & 0.172 & 7.70 & 0.440 & 0.0054 \\
\midrule
\multirow{2}{*}{RaGAN} & SKITTI & 0.252 & 0.168 & 7.85 & 0.598 & 0.0053 \\
 & VSCAPE & 0.232 & 0.133 & 8.83 & 0.508 & 0.0053 \\
\midrule
\multirow{2}{*}{FlwMtch} & SKITTI & 0.274 & 0.187 & 7.52 & 0.555 & 0.0198 \\
 & VSCAPE & 0.24 & 0.129 & 9.01 & 0.504 & 0.0198 \\
\midrule
\multirow{2}{*}{DDPM} & SKITTI & 0.241 & 0.152 & 8.28 & 0.633 & 0.0175 \\
 & VSCAPE & 0.22 & 0.113 & 9.56 & 0.559 & 0.0175 \\
\midrule
\multirow{2}{*}{IUNet} & SKITTI & 0.261 & 0.176 & 7.67 & 0.554 & 0.0122 \\
 & VSCAPE & 0.219 & 0.11 & 9.66 & 0.557 & 0.0122 \\
\midrule
\multirow{2}{*}{M\&M} & SKITTI & 0.282 & 0.197 & 7.19 & 0.506 & 0.0096 \\
 & VSCAPE & 0.228 & 0.116 & 9.43 & 0.544 & 0.0096 \\
\bottomrule
\end{tabular}
\end{center}
\end{table}

\begin{figure}[h]
\begin{center}
\fbox{%
\includegraphics[width=0.8\linewidth]{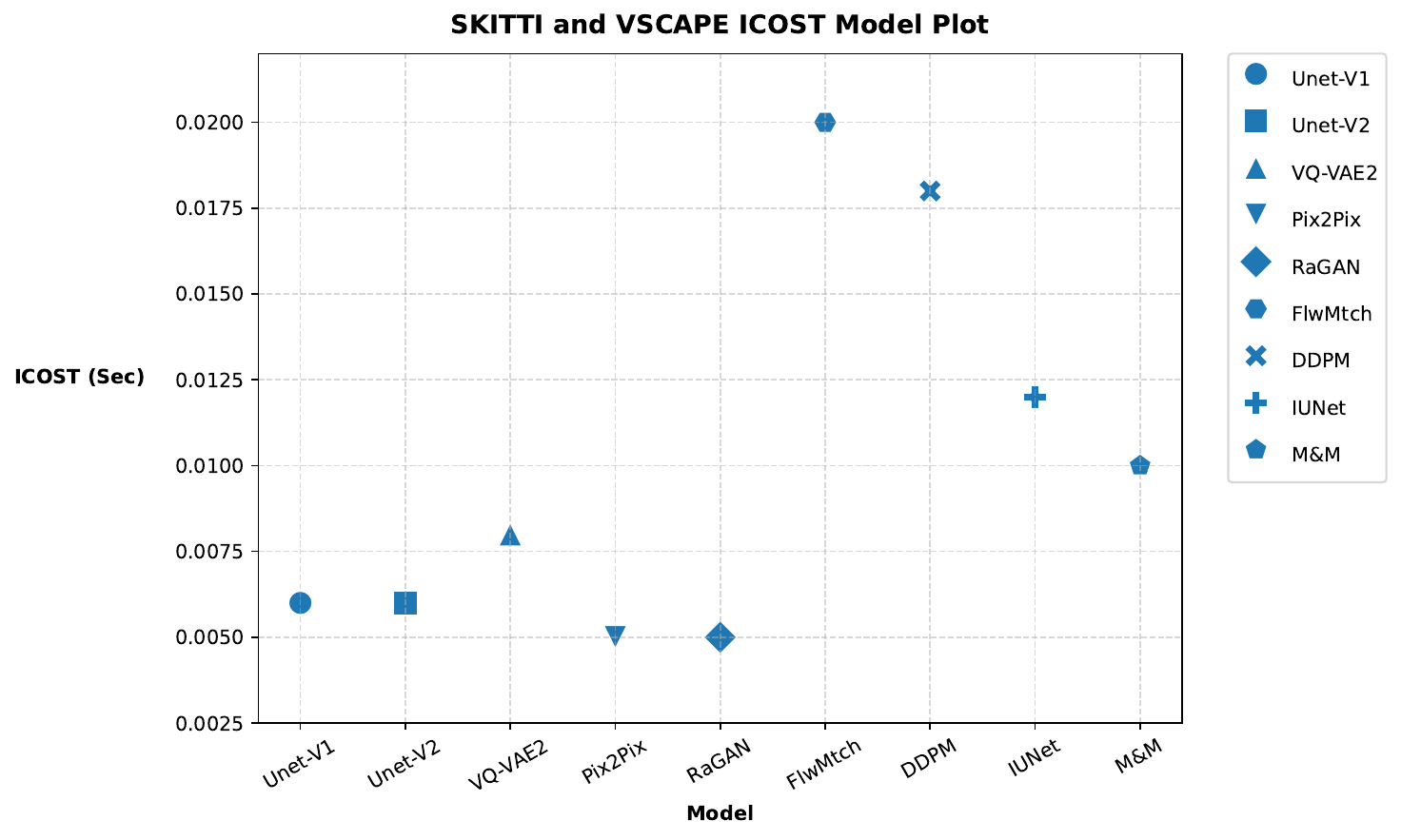}%
}
\end{center}
\caption{SKITTI and VSCAPE ICOST.}
\label{fig:app_icost}
\end{figure}

\FloatBarrier
\begin{equation}
    T_{\mathrm{cost}} = \frac{T_{\mathrm{stop}} - T_{\mathrm{start}}}{B_{\mathrm{size}}} \label{eq:icost}
\end{equation}
where $T_{\mathrm{start}}$ and $T_{\mathrm{stop}}$ denote the recorded start and stop times and $B_{\mathrm{size}}$ denotes the batch-size normalization stated in the original protocol. If the recorded interval spans the entire test dataset rather than a single batch, dividing by batch size does not yield per-frame latency. Because the available description does not fully resolve the timer scope, ICOST is treated as an implementation-specific cost measure rather than independently verified absolute per-frame latency. The relative measurements in Table~\ref{performance-table} are retained unchanged. Figure~\ref{fig:app_icost} shows lower reported ICOST values for UNet-V1, UNet-V2, Pix2Pix, and RaGAN and higher values for the generative DDPM and flow-matching implementations.

\subsection{Sensitivity to the Training Domain}

\begin{figure}[h]
\begin{center}
\fbox{%
\includegraphics[width=0.95\linewidth]{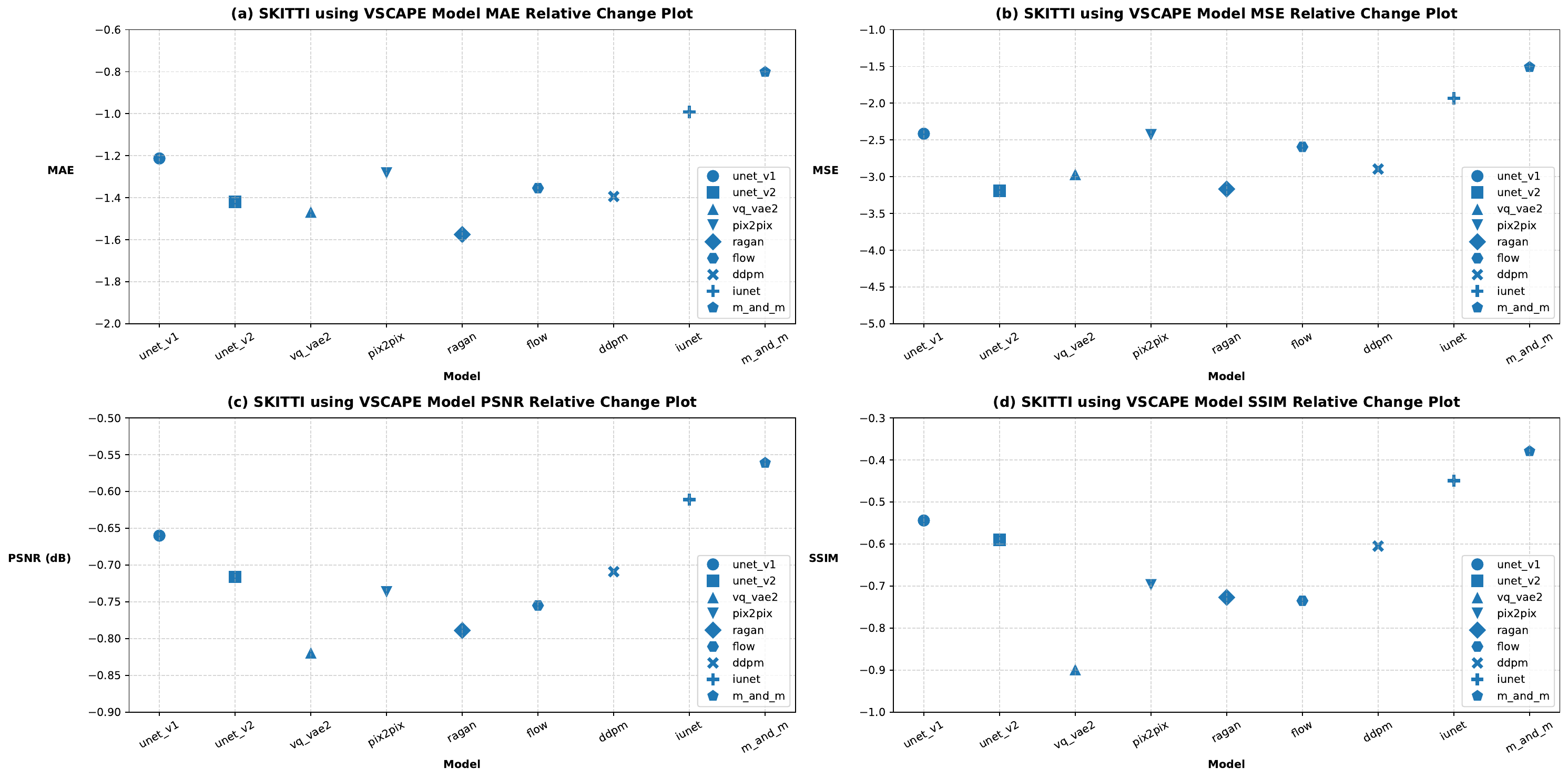}%
}
\end{center}
\caption{Signed relative differences between models trained on VoxelScape and on SemanticKITTI, both evaluated on the SemanticKITTI test set.}
\label{fig:app_rc_plots}
\end{figure}

Figure~\ref{fig:app_rc_plots} compares models trained entirely on SemanticKITTI with models trained entirely on VoxelScape, with both types evaluated on the held-out real SemanticKITTI test set. For each metric, Equation~\ref{eq:relchange} defines a signed relative difference, with the sign chosen so that negative values denote worse synthetic-trained performance. Panels (a)--(d) report MAE, MSE, PSNR, and SSIM, respectively. IUNet and M\&M exhibit large relative differences on several metrics, suggesting sensitivity to the choice of training domain; this does not imply that these models achieve the best cross-domain accuracy.

\FloatBarrier
\begin{equation}
    \mathrm{RC} = s_m\,\frac{M_S - M_V}{M_S} \label{eq:relchange}
\end{equation}
where $M_S$ and $M_V$ are the test metrics obtained using SemanticKITTI-trained and VoxelScape-trained models, respectively, and $s_m=+1$ for MAE and MSE and $s_m=-1$ for PSNR and SSIM. Equation~\ref{eq:relchange} is dimensionless; values can be expressed as percentages only after multiplication by 100.

\subsection{Mixed-Source versus Real-Only Training}

\begin{figure}[h]
\begin{center}
\fbox{%
\includegraphics[width=0.95\linewidth]{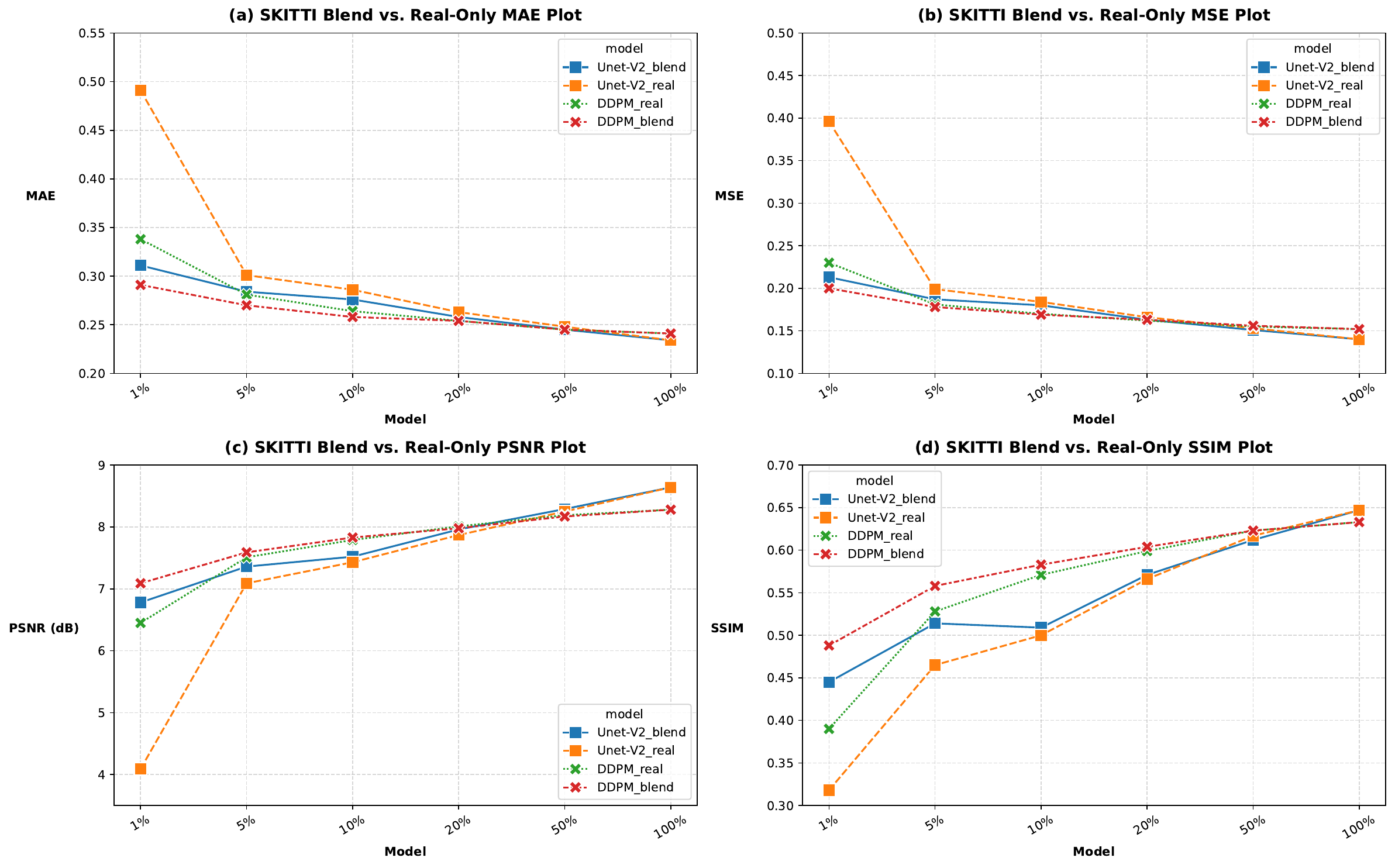}%
}
\end{center}
\caption{Mixed-source versus real-only training on the SemanticKITTI test set; total training-set sizes differ at low real-data fractions.}
\label{fig:app_real_only}
\end{figure}

Figure~\ref{fig:app_real_only} compares the selected mixed-source settings with real-only training for UNet-V2 and DDPM. At 1\% real data, each blend contains 75 real training frames plus 7,425 synthetic frames and a corresponding 25 real validation frames within the 2,500-frame validation mixture. The real-only comparator uses the same available 75 real training frames without the additional synthetic frames. Consequently, the two conditions do not have the same total number of examples or example exposures under a 20-epoch cap. Differences in MAE, MSE, PSNR, and SSIM are shown for completeness, but cannot be attributed solely to the synthetic data without a matched-update control.

\subsection{Variability over Three Random Seeds}

\begin{figure}[h]
\begin{center}
\fbox{%
\includegraphics[width=0.95\linewidth]{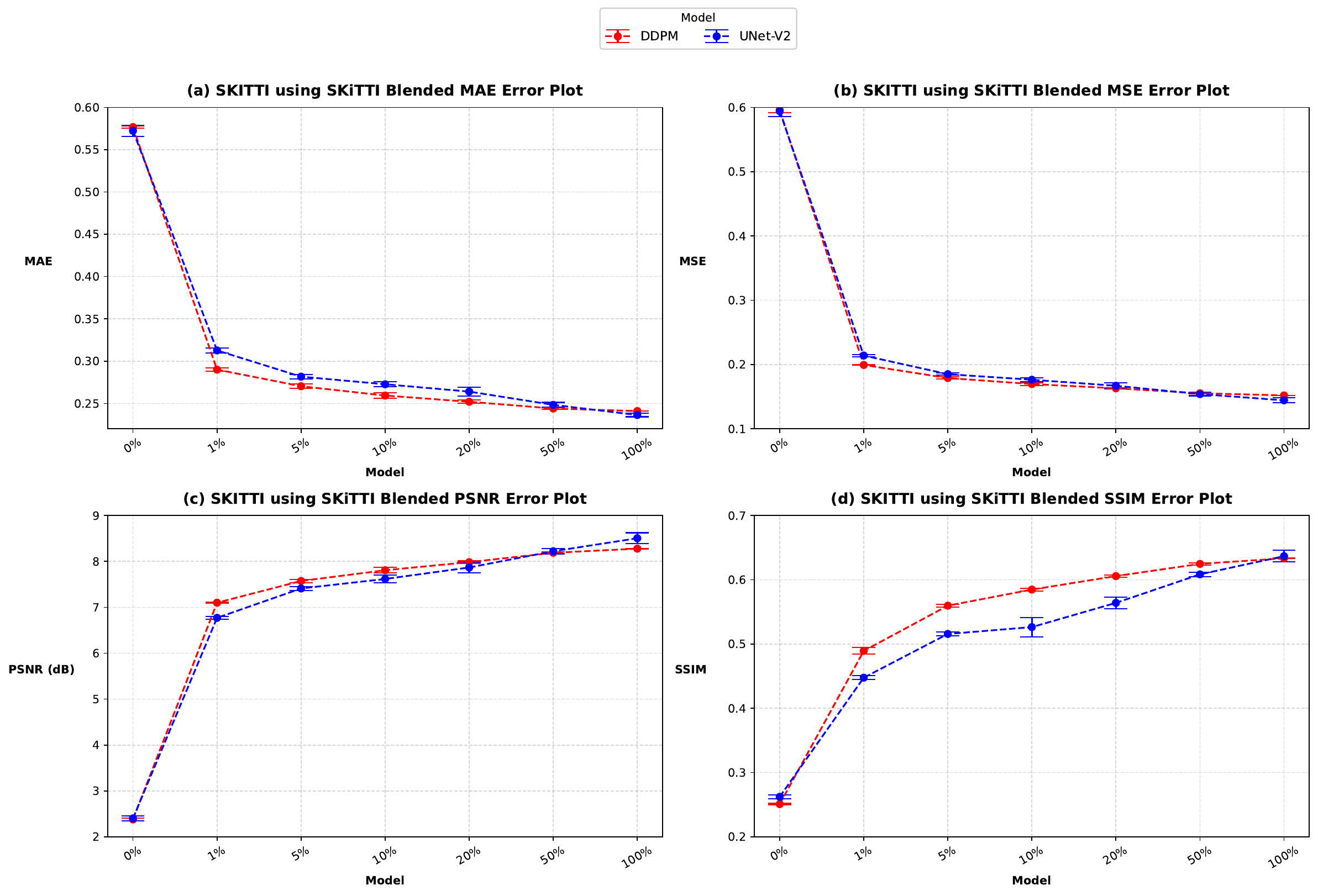}%
}
\end{center}
\caption{Reported metric means and variability for UNet-V2 and DDPM across three random seeds and real-data mixture fractions.}
\label{fig:app_error}
\end{figure}

Figure~\ref{fig:app_error} reports variability across three random seeds (42, 0, and 1337) for DDPM and UNet-V2 at the selected mixture proportions. Panels (a)--(d) display MAE, MSE, PSNR, and SSIM, respectively. These measurements illustrate sensitivity to the reported random seeds, but three runs provide only limited information about the uncertainty of architecture rankings, particularly when the subset of real frames is very small. The original report does not fully specify the error-bar convention (for example, standard deviation or standard error), limiting quantitative uncertainty interpretation.

\subsection{Spectral-Energy Error by Architecture}
\label{section:skitti_blend_spec_err_model}

\begin{figure}[h]
\begin{center}
\fbox{%
\includegraphics[width=0.95\linewidth]{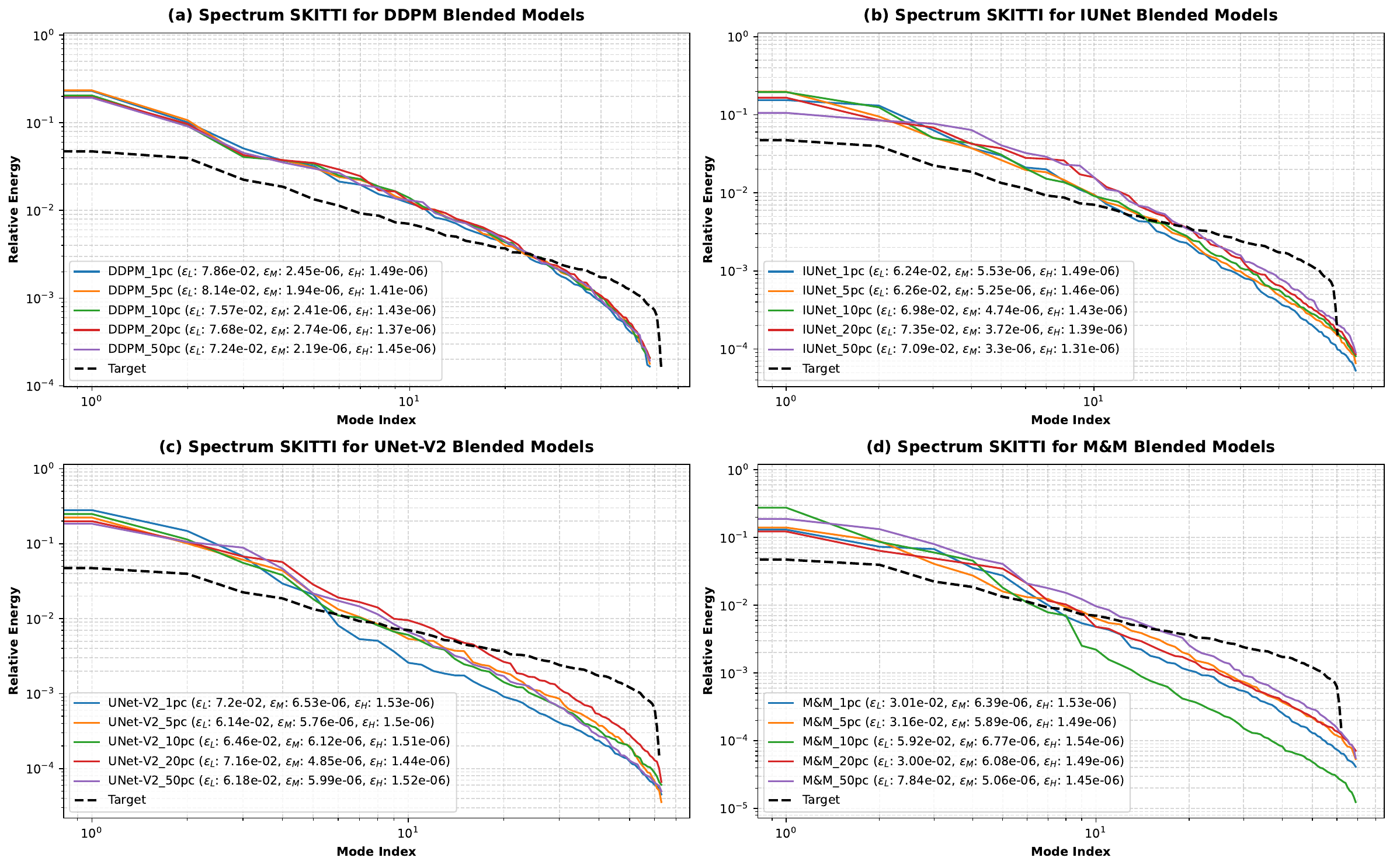}%
}
\end{center}
\caption{Bandwise spectral-energy error for each selected architecture across real/synthetic mixtures.}
\label{fig:app_blend_spec_ene_err}
\end{figure}

Figure~\ref{fig:app_blend_spec_ene_err} summarizes bandwise spectral-energy discrepancies for UNet-V2, DDPM, IUNet, and M\&M across mixture proportions. The reported frequencies are divided into low, medium, and high bands, each containing approximately 21 frequency indices. Equation~\ref{eq:spec_ene_err}, adapted from the spectral-error analysis cited in \citep{khodarkarami2026}, measures the root-mean-square difference between target and predicted squared Fourier magnitudes within a band. Lower values indicate closer energy alignment within that band, not necessarily more accurate spatial locations or object boundaries.

\begin{equation}
    \varepsilon_{\mathcal B} = \sqrt{\frac{1}{|\mathcal B|}\sum_{k\in\mathcal B}\left(\lvert\mathcal F(L)(k)\rvert^2-\lvert\mathcal F(\hat L)(k)\rvert^2\right)^2} \label{eq:spec_ene_err}
\end{equation}
where $L$ and $\hat L$ are the target and predicted intensity images, $\mathcal F$ denotes the Fourier transformation used in the implementation, and $\mathcal B$ is a set of frequency indices belonging to the low, medium, or high band. The Fourier normalization, detailed construction of the frequency bands, and treatment of invalid range-image pixels are not fully documented by the available description. Spectral power does not contain phase information and is therefore a complementary diagnostic rather than a stand-alone measure of image reconstruction quality.

\subsection{Spectral-Energy Error by Mixture Fraction}

\begin{figure}[h]
\begin{center}
\fbox{%
\includegraphics[width=0.95\linewidth]{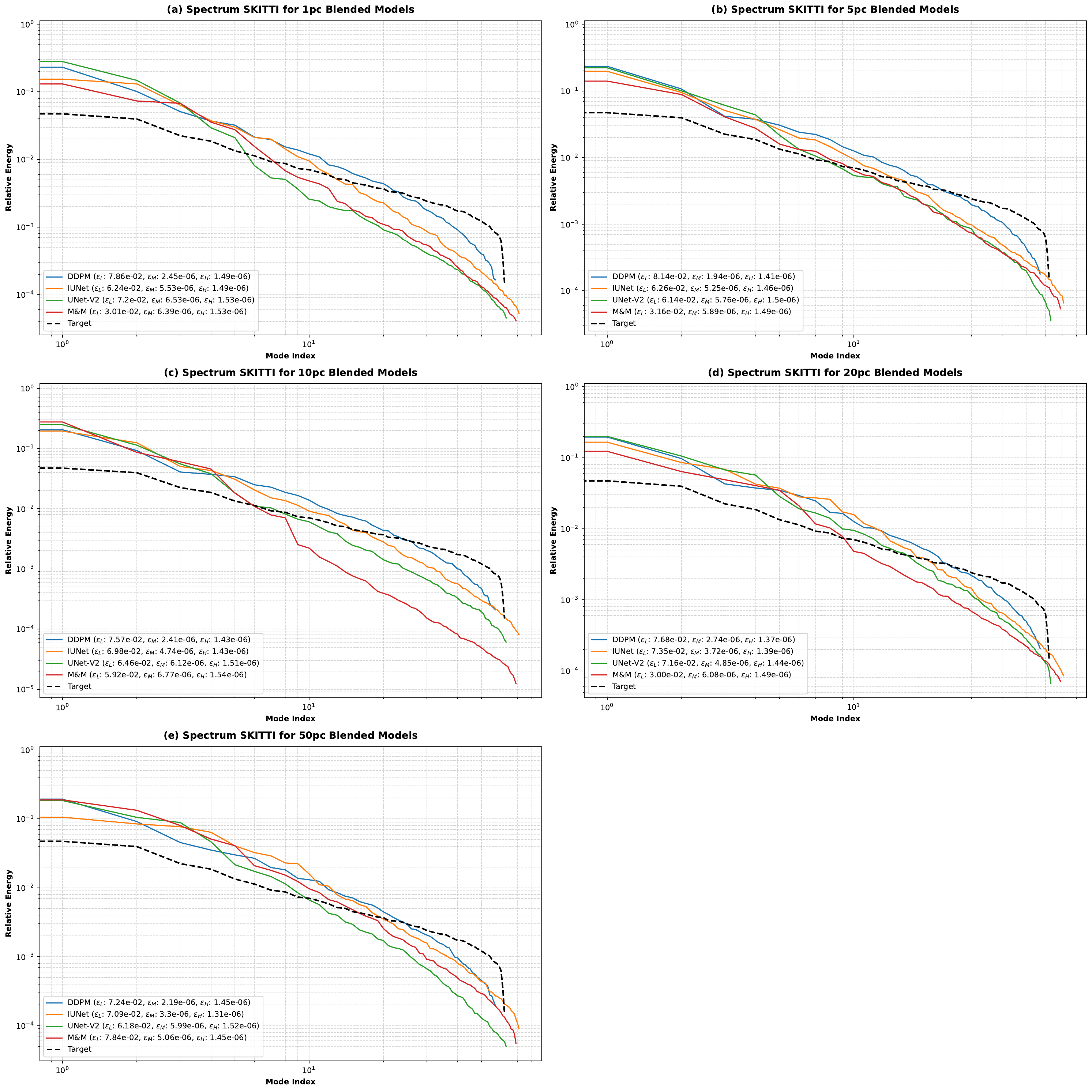}%
}
\end{center}
\caption{Bandwise spectral-energy error for the selected architectures grouped by real-data mixture fraction.}
\label{fig:app_blend_spec_ene_err_pc}
\end{figure}

Figure~\ref{fig:app_blend_spec_ene_err_pc} presents the same low-, medium-, and high-frequency spectral-energy discrepancies grouped by real-data mixture fraction rather than by architecture. These comparisons are descriptive: lower bandwise power error does not necessarily imply lower pixelwise error, sharper boundaries, or correct spatial phase.

\end{document}